# Generalized Qualitative Probability: Savage revisited

**Daniel Lehmann**
Institute of Computer Science
Hebrew University
Jerusalem 91904, Israel
e-mail: lehmann@cs.huji.ac.il

## Abstract

Preferences among acts are analyzed in the style of L. Savage, but as partially ordered. The rationality postulates considered are weaker than Savage's on three counts. The Sure Thing Principle is derived in this setting. The postulates are shown to lead to a characterization of generalized qualitative probability that includes and blends both traditional qualitative probability and the ranked structures used in logical approaches.

## 1  INTRODUCTION

In [Savage, 1954], Savage showed that a set of postulates concerning rational decisions in the face of uncertainty implies that the decider acts as if he/she were maximizing the expectation of some utility function. His postulates imply that a decider gives subjective probabilities, that obey the laws of the calculus of probabilities, to each event. The impact of his results have been felt very strongly and have strengthened the idea that rationality is maximization of expected utility and that subjective probabilities obey the laws of the calculus of probabilities. A number of researchers, though, have contested each of those two points: A. Wald, prior to Savage, has defended a *maximin* strategy irreconcilable with maximization of expected utility, and the Dempster-Shafer approach, among many others, rejects the idea that subjective probabilities are probabilities. In this paper, an approach closely following Savage's is pursued. Some of his assumptions are weakened and an exact characterization of subjective probabilities is obtained.

Three main criticisms may be raised against three of Savage's postulates.

- In P1, Savage [Savage, 1954] requires preference to be a *simple order* (which is a misnomer since, contrary to expectations, it does not imply it is an order relation). He recognizes that this a very strong assumption and writes "There is some temptation to explore the possibilities of analyzing preference among acts as a *partial ordering*, that is, in effect to replace part 1 of the definition of simple ordering by the very weak proposition $f \leq f$, admitting that some pairs of acts are incomparable. This would seem to give expression to introspective sensations of indecision or vacillation, which we may be reluctant to identify with indifference. My own conjecture is that it would prove a blind alley losing much in power and advancing little, if at all, in realism; but only an enthusiastic exploration could shed real light on the question." This work is a first step in this possibly blind alley. *Further work is needed on the utility theory for generalized qualitative probability structures.*

This assumption implies that, if a decider is undecided between two acts $f$ and $g$, and also undecided between $g$ and $h$, then he/she is undecided between $f$ and $h$. In many realistic situations, in which the decider has only partial information, this property cannot be expected to hold. Consider, for example, a choice between betting *heads* in one of three different coin tosses. Coin 1 is believed to be fair. Coins 2 and 3 are unknown, i.e., not believed to be fair, but coin 2 is known to land *heads* more often than coin 3. In such a situation a decider may well be undecided between coins 1 and 2 and also between coins 1 and 3, but definitely prefer coin 2 to coin 3. A criticism of the same nature may be found in [Aumann, 1962], although framed in the von Neumann-Morgenstern setting [von Neumann and Morgenstern, 1947].

There are two different reasons one may object to this *completeness assumption*. The first one, behind [Aumann, 1962], is that one may doubt that consequences may be totally pre-ordered, mainly because consequences may be multi-dimensional (e.g., time and money, human lives and money, or just a very high dimensional space) and one may be either reluctant to trade one dimension for the other or just incapable of describing a total pre-order in a very rich space. Such a concern is at the basis of a large body of work on decision in a many-criteria environment.

The second reason is that one may doubt that subjective probabilities may be exactly described (i.e., events totally pre-ordered), because of one's uncertainty about what is unknown, or because one may be reluctant to compare subjective probabilities of events pertaining to very different realms, e.g., the probability of a Republican president being elected in 1996 (Savage's example) and that of a coin landing "heads". Such a concern is at the basis of a large body of work on decision in uncertain environments.

The results of this paper do not depend on the source of incompleteness. It has been elaborated with the second point of view in mind, though.

• Savage's treatment of null events is disputable since it does not allow for events that are null relatively to other null events. As any treatment based on probabilities, or reducible to them, it allows conditioning on an event $H$ only if the probability of $H$ is positive. This point is made in [Blume *et al.*, 1991] where *non-archimedean* probability structures are proposed, in the framework of a *total* pre-order. There, reference is made to [Blackwell and Dubins, 1975] and de Finetti [de Finetti, 1972] is quoted as saying: "there seems to be no justification ... for introducing the restriction $P(H) \neq 0$".

• It follows from Savage's postulate P4 that if a consequence $c$ is strictly preferred to a consequence $d$, and if the act of winning $c$ in case of an event $A$ and winning $d$ otherwise is strictly preferred to winning $c$ in case of an event $B$ and winning $d$ otherwise, then the act of winning $c'$ in case of $A$ and winning $d'$ otherwise is strictly preferred to winning $c'$ in case of $B$ and winning $d'$ otherwise, for any consequences $c'$, $d'$ such that $c'$ is strictly preferred to $d'$. This seems too strong. One may, rationally, strictly prefer winning \$1M on $A$ and winning \$0 otherwise to winning \$1M on $B$ and winning \$0 otherwise, be undecided between winning 1c on $A$ and \$0 otherwise or winning 1c on $B$ and \$0 otherwise, and still prefer a sure win of 1c to nothing. The treatment of subjective probabilities in [Anscombe and Aumann, 1963], similarly, includes no analogue to P4. Note Savage's pessimistic outlook in [Savage, 1954, p. 30, Section 2]: "Though I have not explored the latter possibility carefully, I suspect that any attempt to do so formally leads to fruitless and endless regression".

This work presents postulates, in the style of Savage, that are weak enough to answer all three criticisms above. Nevertheless, those postulates enable a full characterization of the structure of *subjective probabilities*.

## 2   DECISION IN THE FACE OF UNCERTAINTY

Notations follow those of [Savage, 1954]. $S$ is a non-empty set of states. Subsets of $S$ are noted $A$, $B$, $C$, ... and called events. We follow Savage in assuming that every subset of $S$ is an event, i.e., measurable. The treatment of the general case, in which the measurable subsets of $S$ form only a sub-algebra of the power set of $S$ poses no serious problems. $F$ is a set of consequences, denoted $c$, $d$, .... Acts are arbitrary functions from $S$ to $F$, denoted by $f$, $g$, $h$, ....

A basic problem with Savage's treatment is the elimination of the event of reference, i.e., the event on which the decider conditions, from the formalization. The preference relation shall here be indexed by an arbitrary event $A$. For every event $A$: $f \leq_A g$ means that, given $A$, *either $g$ is strictly preferred to $f$ or one is indifferent between $f$ and $g$*. The reason for this richer formalism is *not* a rejection of the Sure Thing Principle, which is accepted and derived, but the rejection of Savage's idea that comparing two acts that agree on $A$ is enough to compare them given $A$. If $A$ is negligible with respect to $\bar{A}$, acts may be equivalent (on the whole set $S$) but not given $A$.

Notice also that the intuitive meaning of $\leq$ is *preference or indifference*, as in [Aumann, 1962], and *not* non-preference as in [Savage, 1954]. In this way, any two acts may stand in four possible situations:

• $f \leq_A g$ and $g \not\leq_A f$, i.e., $g$ is strictly preferred to $f$, given $A$, which will be denoted $f <_A g$,

• $g \leq_A f$ and $f \not\leq_A g$, i.e., $f$ is strictly preferred to $g$, given $A$, which will be denoted $g <_A f$,

• $f \leq_A g$ and $g \leq_A f$, i.e., $f$ and $g$ are indifferent, given $A$, which will be denoted $f \sim_A g$, or

• $f \not\leq_A g$ and $g \not\leq_A f$, i.e., one is undecided between $f$ and $g$, given $A$.

We propose, in this work, two weakenings of Savage's postulates: a partial pre-order in place of a total pre-order, and the consideration of events that are negligible with respect to other events (that may be negligible with respect to a third class of events). Those two weakenings are essentially orthogonal, and one may study either one of them in isolation. The postulates proposed will be presented and discussed now. The first two postulates require the preference relation, on any event $A$, to be a pre-order.

(**Q0**)   $f \leq_A f$.

(**Q1**)   $f \leq_A g$, $g \leq_A h$ $\Rightarrow$ $f \leq_A h$.

Notice that the relation $\leq_A$ does not satisfy the first property of the definition of a simple ordering on p.18 of [Savage, 1954], i.e., either $f \leq_A g$ or $g \leq_A f$, but satisfies the second property, i.e., transitivity. Postulates (Q0) and (Q1) are therefore strictly weaker than Savage's (P1). It clearly follows from (Q0) and (Q1) that, on any event $A$, $<_A$ is irreflexive and transitive, i.e., a strict partial order, and that $\sim_A$ is reflexive, symmetric and transitive, i.e., an equivalence relation. Notice that the event $A$ may be empty: $f$ is not strictly preferred to $f$, even if the empty event obtains, but one is indifferent between $f$ and itself even if the empty event obtains. It also follows from (Q0) and (Q1) that indifferent acts behave in exactly the same way concerning the preference relation, i.e., if $f \sim_A g$ then

$$f \sim_A g,\ h <_A f,\ f <_A k\ \Rightarrow\ h <_A g,\ g <_A k.$$

So far, all postulates considered one fixed event $A$. The next postulates deal with the influence of the event $A$ in the relation $<_A$. First, preferences on $A$ should not depend on the values of the acts $f$ and $g$ outside $A$.

**Definition 1** *Acts $f$, $g$ are said to be equivalent on event $A$ iff for every $s \in A$, $f(s) = g(s)$. This will be denoted by $f =_A g$.*

Our postulate says only that, given $A$, one should be indifferent between acts that are equivalent on $A$.

(Q2)    $f =_A g\ \Rightarrow\ f \sim_A g.$

Notice that (Q2) implies (Q0). A consequence of (Q2) is that equivalent acts are indeed equivalent with respect to the preference relation.

**Lemma 1** *If $f =_A f'$ and $g =_A g'$, then $f \leq_A g$ implies $f' \leq_A g'$.*

Another consequence is that the preference order on the empty event is trivial.

**Lemma 2** *$\forall h, h',\ h \leq_\emptyset h'$ and therefore $h \not<_\emptyset h'$.*

Events different from the empty event may yield a trivial relation: such events will be called null events.

**Definition 2** *An event $A$ is null iff, $\forall h, h',\ h \leq_A h'$.*

Definition 2 is similar in spirit with Savage's definition of null events, but technically different, since the formalism used here is richer and we may conditionalize explicitly on the event $A$. The next two postulates consider two disjoint events $A$ and $B$ (this assumption is essential), and two acts $f$ and $g$ that are indifferent given $B$. Postulate (Q3) assumes information on the preferences on $A$ to deduce information concerning the preferences on $A \cup B$. Postulate (Q4) assumes information on $A \cup B$ and deduces information on $A$.

(Q3)    $A \cap B = \emptyset,\ f \leq_A g, f \sim_B g\ \Rightarrow\ f \leq_{A \cup B} g$

If one is indifferent between $f$ and $g$ given $B$ and either indifferent or prefers $g$ given $A$, one should be indifferent between them given $A \cup B$ or prefer $g$, one could not reasonably prefer $f$ or even be undecided.

The condition $A \cap B = \emptyset$ is essential. If it were not satisfied, it could be the case that one prefers $g$ to $f$ on the intersection of $A$ and $B$, $f$ being preferred to $g$ on the symmetric difference. In such a case, it may well be the case that $f$ is overall preferred to $g$ on the union, but $g$ is preferred to $f$ on both $A$ and $B$.

**Corollary 1** *If $A \cap B = \emptyset$, $f \sim_A g$ and $f \sim_B g$, then $f \sim_{A \cup B} g$.*

Some notation will be helpful for our next postulate. We shall say that $A$ is negligible compared to $B$ if $A$ and $B$ are disjoint and the relations of indifference and preference on the union $A \cup B$ are exactly as on $B$.

**Definition 3** *Assume $A \cap B = \emptyset$. We shall say that $A$ is negligible compared to $B$ and write $A \mathcal{N} B$ if $\forall h, h',\ h \leq_{A \cup B} h' \Leftrightarrow h \leq_B h'$.*

Notice that $A \mathcal{N} B$ implies that $\forall h, h',\ h \sim_{A \cup B} h'$ iff $h \sim_B h'$ and $h <_{A \cup B} h'$ iff $h <_B h'$. We may now express our next postulate. Suppose $A$ and $B$ are disjoint. Suppose also that, on the union $A \cup B$, $g$ and $f$ are indifferent or $g$ is preferred, and that, on $B$, $g$ and $f$ are indifferent. Then two cases may arise: either $f \leq_A g$, explaining, in accordance with (Q3) that $f \leq_{A \cup B} g$, or $A$ is negligible with respect to $B$.

(Q4)    $A \cap B = \emptyset,\ f \leq_{A \cup B} g,\ f \sim_B g\ \Rightarrow$

either $f \leq_A g$ or $A \mathcal{N} B$.

**Corollary 2** *If $A \cap B = \emptyset$, $f \sim_{A \cup B} g$ and $f \sim_B g$, then either $f \sim_A g$ or $A \mathcal{N} B$.*

Our next corollary says that if $g$ is strictly preferred to $f$ on $A$, but $f$ and $g$ are indifferent on $B$, then two cases may occur: either $A$ is not negligible w.r.t. $B$ and $g$ is strictly preferred to $f$ on the union $A \cup B$, or, $A$ is negligible w.r.t. $B$, and therefore $f$ and $g$ are indifferent on $A \cup B$.

**Corollary 3** *If $A \cap B = \emptyset$, $f <_A g$ and $f \sim_B g$, then either $f <_{A \cup B} g$ or $A \mathcal{N} B$. Notice that, in the second case, $f \sim_{A \cup B} g$.*

If $g$ is strictly preferred to $f$ on the union, but $f$ and $g$ are indifferent on $B$, $g$ must be preferred to $f$ on $A$.

**Corollary 4** *If $A \cap B = \emptyset$, $f <_{A \cup B} g$ and $f \sim_B g$, then $f <_A g$.*

**Lemma 3** *Assume $A \cap B = \emptyset$.    If $f \leq_A g$ and $f \leq_B g$, then $f \leq_{A \cup B} g$.*

**Proof:** Let

$$h(s) = \begin{cases} g(s) \text{ if } s \in A \\ f(s) \text{ otherwise.} \end{cases} \tag{1}$$

We have, by (Q2), $h \sim_A g$ and $h \sim_B f$. Therefore $f \leq_A h$ and $f \sim_B h$. By (Q3), then $f \leq_{A \cup B} h$. Similarly, $h \sim_A g$ and $h \leq_B g$ and, therefore, $h \leq_{A \cup B} g$. By (Q1), then, $f \leq_{A \cup B} g$. ∎

One may now prove a result similar to Lemma 3 for strict preferences.

**Lemma 4** *Assume* $A \cap B = \emptyset$. *If* $f <_A g$ *and* $f <_B g$, *then* $f <_{A \cup B} g$.

A property stronger than postulate (Q4) may be considered by not allowing the second possibility in the conclusion:

$(\mathbf{Q'4})$ $A \cap B = \emptyset$, $f \leq_{A \cup B} g$, $f \sim_B g \Rightarrow f \leq_A g$.

This property is consistent with the postulates since it is satisfied by models in which acts are compared by their expectations. We shall now show that the postulates above imply Savage's Sure Thing Principle (Postulate P2).

**Lemma 5** *Let* $A \cap B = \emptyset$. *If* $f =_A f'$, $g =_A g'$, $f =_B g$ *and* $f' =_B g'$, *then* $f \leq_{A \cup B} g$ *implies* $f' \leq_{A \cup B} g'$.

**Proof:** By (Q2), we have $f \sim_A f'$, $g \sim_A g'$, $f \sim_B g$ and $f' \sim_B g'$. If $A \mathcal{N} B$, then $f' \sim_B g'$ implies $f' \sim_{A \cup B} g'$. We may, therefore, without loss of generality, assume that $A$ is not negligible with respect to $B$. Since $f \leq_{A \cup B} g$ and $f \sim_B g$, (Q4) then implies, since $A$ is not negligible with respect to $B$, that $f \leq_A g$. Therefore $f' \leq_A g'$ and, by (Q3), $f' \leq_{A \cup B} g'$. ∎

The next two postulates deal with the constant acts, i.e., with the preference ordering on consequences.

**Definition 4** *An act* $f$ *is constant iff for any* $s, t \in S$, $f(s) = f(t)$.

A constant act $f$ may be identified with the consequence $f(s)$ for $s \in S$. Preferences on constant acts are independent of the event considered.

$(\mathbf{Q5})$ If $c, d$ are constants and $A$ is not null,

and $c \leq_A d$, then for any event $B$, $c \leq_B d$.

Postulate (Q5) is a slight strengthening of Savage's (P3), since, in the latter, the condition on $A$ is that $A$ be not null, whereas (Q5) uses the stronger condition $A$ non-empty. This strengthening is not significant for two reasons:

- we could use a weaker version of (Q5), introducing the notion of a null event as in Savage, and
- null events may be treated as events of positive probability infinitesimally close to zero.

From here on, preference between constants will be denoted $\leq$ without subscript. The restriction that $A$

be non-null is clearly crucial since, otherwise, for any $h, h', h \leq_A h'$.

Our last postulate ensures non-triviality: there are two constants, i.e., consequences, one of them preferred to the other.

$(\mathbf{Q6})$ There are constants $c, d$ such that $d < c$.

Notice that we do not assume the set $F$ of consequences is totally or even modularly ordered (i.e., $d < c$ implies that, for any $e$, either $d < e$ or $e < c$). Notice also that we have no postulate similar to Savage's (P4). The next section will show that any system of preferences and indifferences that satisfies (Q1)–(Q6) yields a binary relation on events that enables us to compare the (generalized) probability of events. This relation enjoys extremely interesting properties.

# 3 GENERALIZED QUALITATIVE PROBABILITY

Postulates (Q5) and (Q6) deal with constant acts, i.e., acts that take only one single value. Of special importance are also acts that take only two different values. Suppose $c$ and $d$ are different consequences and that $f$ is an act that takes only values $c$ and $d$, i.e., on some event $A$, $f$ takes value $c$ and on the complement of $A$, $\bar{A}$, it takes value $d$. Let us devise the following notation:

$$w_A^{c,d}(s) = \begin{cases} c \text{ if } s \in A \\ d \text{ otherwise.} \end{cases} \quad (2)$$

Assuming $d < c$, $w_A^{c,d}$ "wins" on $A$, i.e., gets on $A$ the high pay-off $c$ and "loses", i.e., gets the low pay-off $d$ on $\bar{A}$. The following may be shown.

**Lemma 6** *If* $A \cup B \subseteq D$, $w_A^{c,d} \leq_{A \cup B} w_B^{c,d}$ *implies* $w_A^{c,d} \leq_D w_B^{c,d}$.

The meaning of $w_A^{c,d} \leq_{A \cup B} w_B^{c,d}$ is: winning on $B$ is preferred to or indifferent with winning on $A$. Could it be that preferences depend on the prizes $c$ and $d$? In many circumstances, probably not, and Savage has a postulate, (P4), on p. 31 to that effect. In the present framework, we may easily formalize this by: if $d < c$ and $d' < c'$, then

$(\mathbf{R})$ $w_A^{c,d} \leq_{A \cup B} w_B^{c,d} \Rightarrow w_A^{c',d'} \leq_{A \cup B} w_B^{c',d'}$

A restricted version, in which $A$ and $B$ are assumed to be disjoint, is equivalent. A difference between (R) and (P4) should be noted: (R) is relativized to $A \cup B$, as is our constant policy.

We shall, after some discussion, reject this postulate, but let us see, first, the argument in its favor. If one prefers to win on $B$ than to win on $A$, this should not depend on the prizes offered: if two horses are at equal odds and one prefers to wage \$1,000 on one of them, "archie" than on the other one , "belle", then one must also prefer to wage \$1 on "archie" than on "belle". The reason is clearly that one must think that the chances of "archie" winning are better than those of "belle" winning, and this enough to convince us to strictly prefer waging even a small sum on "archie" than on "belle", at least if the small sum is big enough to be strictly preferred to \$0.

Now, the argument against (R). Suppose $A$ is an event, e.g., the result of a lottery, for the probability of which one has a precise and reliable estimation. Suppose, on the contrary, one has only a very fuzzy estimation of the probability of $B$. A rational decider may well, if the sum involved is small, prefer to bet on $B$ than on $A$, but prefer to bet on $A$ if the sum involved is large. Similarly, the choice of $A$ or $B$ may depend on whether a gain or a loss is expected.

As explained above, we do not accept (R). Given a system of preferences and indifferences, one may naturally define a relation on events.

**Definition 5** *We shall say that $A$ is not more plausible than $B$, and write $A \leq B$ iff $\forall c, d$, such that $d < c$, one has $w_A^{c,d} \leq_{A \cup B} w_B^{c,d}$. The relation $\leq$ on events will be said to be defined by the preference structure $\leq_-$ on acts.*

The meaning of Definition 5 is he following: if one prefers betting on $B$ than on $A$, whatever the prize is, it must be because one considers $B$ as more probable than $A$.

Let us now consider the consequences of Definition 5. First, one prefers losing on a less probable event than on a more probable event.

**Lemma 7** *If $A \leq B$, then $c \leq d \Rightarrow w_B^{c,d} \leq_{A \cup B} w_A^{c,d}$.*

**Proof:** For $d \sim c$, by (Q5) and (Q2), $w_A^{c,d} \sim_{A-B} w_B^{c,d}$ and $w_A^{c,d} \sim_{B-A} w_B^{c,d}$. Since $w_A^{c,d} =_{A \cap B} w_B^{c,d}$, by (Q2) and Corollary 1, we see that $w_A^{c,d} \sim_{A \cup B} w_B^{c,d}$. For $c < d$, by Definition 5, $w_A^{d,c} \leq_{A \cup B} w_B^{d,c}$. Let $f = w_A^{d,c}$, $g = w_B^{d,c}$, $f' = w_B^{c,d}$ and $g' = w_A^{c,d}$. Let also $D = (A - B) \cup (B - A)$ and $E = A \cap B$. We have: $D \cap E = \emptyset$, $f =_D f'$, $g =_D g'$, $f =_E g$ and $f' =_E g'$. Lemma 5 (The Sure Thing Principle) says that $f \leq_{D \cup E} g$ implies $f' \leq_{D \cup E} g'$. ∎

Before embarking on a study of the properties of the relations $\leq$ on events, one would like to be convinced that the postulates (Q1)–(Q6) are consistent and consider some models for them. Let $F$ be the unit interval and assume $S$ is a probability space, in which every event is measurable. Let any event of probability zero be null, and for $A$ of positive probability, define $f \leq_A g$ iff the expected value of $f$ conditioned on $A$ is less or equal to that of $g$ conditioned on the same event. It is easy to see that all postulates are satisfied, and it is worth noticing that the stronger property (Q'4) is also satisfied. The restriction that the empty event be the only event of probability zero is needed because (Q5) is a slight strengthening of Savage's (P3). Another, perhaps more interesting and more widely applicable, model of the postulates is obtained if one considers a non-standard probability measure on $S$ in which every event is measurable and the empty event is the only event of probability zero. Notice that we may have non-empty events of infinitesimally small probability. Notice also that we take $F$ to be the standard unit interval and do not allow infinitesimally small consequences. For $A$ of positive probability, define $f <_A g$ iff the expected value of $f$ conditioned on $A$ is less than that of $g$ conditioned on the same event, *and the difference is not infinitesimal*. If $A$ has zero probability, it is a null event. A third way of generating preferences is the following. Assume $S$ is totally ordered in a way every event has a maximum. Intuitively $s < t$ means that state $t$ is more plausible than $s$ to such an extent that, given an event that contains both, one should not be influenced by the consequences on $s$. Given an event $A$, let $s_A$ be the maximal element of $A$. Given two acts $f$ and $g$ and an event $A$, we shall compare $f$ and $g$ on $A$ by the values they take on $s_A$: $f \leq_A g$ iff $f(s_A) \leq g(s_A)$.

We are now going to prove properties of the relation $\leq$ on events. They parallel the definition of qualitative probability given by Savage in [Savage, 1954, page 32] and will be used as the definition of *generalized qualitative probabilities* given in Definition 6. Our first lemma states that $\leq$ is a pre-order.

**Lemma 8** *The relation $\leq$ on events is reflexive and transitive.*

As usual we shall write

- $A \sim B$ for $A \leq B$ and $B \leq A$, and
- $A < B$ for $A \leq B$ and $B \not\leq A$.

The relation $\sim$ is an equivalence relation and $<$ is a strict partial order. Lemma 8 generalizes part 1 of the definition of qualitative probability in [Savage, 1954, p. 32]: $\leq$ is a simple ordering. We now prove results that parallel part 2 there: $B \leq C$ iff $B \cup D \leq C \cup D$, provided $B \cap D = C \cap D = \emptyset$. D. Dubois remarked, some time ago, that the two directions implied by the "if and only if" there were not at all equally obvious, or desirable. The "only if" part seems inescapable, and we prove now it holds for generalized qualitative probability.

**Lemma 9** *Let $A \cap D = B \cap D = \emptyset$. If $A \leq B$, then $A \cup D \leq B \cup D$.*

The "if" part does not hold in our framework. A weaker property will be presented in Lemma 12 but, first, the exact counterpart of the first part of part 3 of the definition of qualitative probability. We now prove properties that parallel property 3 there.

**Lemma 10** *For any event $A$, $\emptyset \leq A$.*

**Corollary 5** *If $A \subseteq B$, then $A \leq B$.*

The following will be helpful.

**Lemma 11** *If $A \cap B = \emptyset$, then $A \mathcal{N} B$ iff $(B \cup A) \leq B$.*

**Proof:** Let $A$ and $B$ be disjoint. Suppose, first, that $A \mathcal{N} B$. Assume $d < c$. We must show that $w_{B \cup A}^{c,d} \leq_{B \cup A} w_B^{c,d}$. But this follows from $w_{B \cup A}^{c,d} =_B w_B^{c,d}$.

Assume, now, that $(B \cup A) \leq B$. By (Q6) there are constants, $c$ and $d$, such that $d < c$. We have $w_{B \cup A}^{c,d} \leq_{B \cup A} w_B^{c,d}$. But $w_{B \cup A}^{c,d} =_B w_B^{c,d}$, and (Q2) and (Q4) imply that either $w_{B \cup A}^{c,d} \leq_A w_B^{c,d}$, or $A \mathcal{N} B$. In the second case, we are through. In the first case, $c \leq_A d$, and, by (Q5), $A$ is null, and therefore $A \mathcal{N} B$. ∎

We may now present the lemma announced above.

**Lemma 12** *Let $A \cap D = B \cap D = \emptyset$. If $A \cup D \leq B \cup D$ and $(D \cup B) \not\leq D$, then $A \leq B$.*

Our last lemma is more original: it is a strengthening of the property $\emptyset < S$ of the definition of subjective probability. It expresses the fact that a sum must be greater than at least one of its parts.

**Lemma 13** *If $A$ and $B$ are disjoint events, $A \leq B$ and $A \cup B \leq A$, then, $A$ and $B$ are null events.*

**Proof:** Let $A \cap B = \emptyset$, $A \leq B$ and $A \cup B \leq A$. Let $d < c$, as guaranteed by (Q6). We know that $w_{A \cup B}^{c,d} \leq_{A \cup B} w_A^{c,d}$, and $w_{A \cup B}^{c,d} \sim_A w_A^{c,d}$. By (Q4), either $w_{A \cup B}^{c,d} \leq_B w_A^{c,d}$, i.e., $c \leq_B d$, implying $B$ is a null event, or $B \mathcal{N} A$. In both cases, then, $w_A^{c,d} \leq_{A \cup B} w_B^{c,d}$ implies $A \cup B$ is a null event. A subset of a null event is a null event. ∎

The following corollary says that a non-null event, and in particular $S$, is strictly larger than the empty set.

**Corollary 6** *An event $A$ is null iff $A \leq \emptyset$.*

We may now encapsulate the properties above in a definition. It strictly generalizes the definition of *qualitative probability*[Savage, 1954, p. 32].

**Definition 6** *A reflexive and transitive relation $\leq$ on the subsets of $S$ is a generalized qualitative probability iff:*

1. *$A \cap D = B \cap D = \emptyset$, $A \leq B \Rightarrow A \cup D \leq B \cup D$,*

2. *$A \cap D = B \cap D = \emptyset$, $A \cup D \leq B \cup D$, $D \cup B \not\leq D \Rightarrow A \leq B$,*

3. *if $A \cap B = \emptyset$, $A \leq B$ and $A \cup B \leq A$, then $B \leq \emptyset$,*

4. *for any event $A$, $\emptyset \leq A$.*

Notice that we do not ask that $A \leq B$ imply $\bar{B} \leq A$. This property does not follow from our requirements. We have shown that, given any preference structure satisfying (Q1)–(Q6), the relation described in Definition 5 is a generalized qualitative probability (g.q.p.).

# 4 PROPERTIES OF GENERALIZED QUALITATIVE PROBABILITY

Let us now consider properties of generalized qualitative probabilities. Most of the properties we shall prove are needed in the proof of Theorem 1, others are of independent interest. For lack of space, proofs will not be given.

**Lemma 14**
- *An event $A$ is null iff $\emptyset \leq A$.*

- *In a g.q.p., $A \subseteq B$ implies $A \leq B$.*

- *$A \cap D = B \cap D = \emptyset$, $A \cup D < B \cup D \Rightarrow A < B$.*

- *$A \cap D = B \cap D = \emptyset$, $A < B$, $D < B \cup D \Rightarrow A \cup D < B \cup D$.*

- *If $A \cap B = \emptyset$, $A' \leq A$ and $B' \leq B$, then $A' \cup B' \leq A \cup B$.*

- *If $A \cap B = \emptyset$ and $\emptyset < A \cup B$, then either $A < A \cup B$ or $B < A \cup B$.*

A fundamental notion is needed to study further the properties of g.q.p.

**Definition 7** *We shall say that $A$ is negligible with respect to $B$, and write $A \ll B$ iff $B \not\leq \emptyset$ and $A \cup B \leq B - A$.*

The intuition is that $A$ is negligible w.r.t. $B$ iff both $A \cap B$ and $A - B$ are negligible w.r.t. $B$, i.e., $B \leq B - (A \cap B)$ and $B \leq B \cup (A - B)$. The properties of the relation $\ll$ are many and delicate to prove. Again no proofs will be given. The main result we need about $\ll$ (all needed properties will easily follow) is that $\ll$ is modular, i.e., if $A \ll C$, then, for any $B$, either $A \ll B$, or $B \ll C$. The term *modular* is taken from Grätzer [Grätzer, 1971]).

**Lemma 15**
- *If $A \subseteq B \ll C$, then $A \ll C$.*

- *If $A \ll B \subseteq C$, then $A \ll C$.*

- *If $A \ll B$, then $A < B$.*

- *Assume $B \subseteq C$. If $A \ll C$, then either $A \ll B$ or $B \ll C$.*

- *Assume $B \cap C = \emptyset$. If $A \ll C$, then either $A \ll B$ or $B \ll C$.*

- *If $A \ll C$, then, for any $B$, either $A \ll B$ or $B \ll C$.*

- If $A \leq B \ll C \leq D$, then $A \ll D$.

- If $A \ll B$ and $A' \ll B$, then $A \cup A' \ll B$.

- If $A \ll B \cup B'$, then either $A \ll B$ or $A \ll B'$.

- If $A \cap D = B \cap D = \emptyset$ and $A \cup D \leq B \cup D$, then either $A \leq B$ or $A \ll B \cup D$.

- If $A \cap D = B \cap D = \emptyset$ and $A \cup D \leq B \cup D$, then either $A \leq B$ or $(A \cup B) \ll D$.

- If $A \cap D = \emptyset$ and $A \cup D \leq B \cup D$, then either $A \leq B$ or $(A \cup B) \ll D$.

- If $A \cap B = \emptyset$, $A \cup B \leq A' \cup B'$ and $B' \leq B$, then either $A \leq A'$ or $(A \cup A') \ll B'$.

It may now been shown that there is no additional property that should be added to the definition of g.q.p.

**Theorem 1** *If $\leq$ is a generalized qualitative probability, there is a preference structure satisfying (Q1)–(Q6) that defines it, in the sense of Definition 5.*

**Proof:** Let $F = \{high, low\}$. Since acts can take only two values, every act is of the form $w_A^{high,low}$ for some event $A$ ($A$ is the set of states on which the act takes the value *high*). We shall drop the upper index and write $w_A$. Let us define:

$$w_A \leq_D w_B \text{ iff } (D \cap A) \ll D \text{ or } A \cap D \leq B \cap D.$$

Notice immediately that $w_A \leq_{A \cup B} w_B$ iff $A \ll B$ or $A \leq B$, which holds iff $A \leq B$. Therefore $\leq$ is the relation defined by the preferences on acts that have been just defined and the only task left to us is to check that postulates (Q1)–(Q6) are satisfied. In fact property (R) is also satisfied. The proof will appear in the full paper. ∎

# 5   FURTHER RESULTS

## 5.1   FAMILIES OF GENERALIZED QUALITATIVE PROBABILITIES

A number of interesting families of generalized qualitative probabilities may be defined.

**Definition 8**   • *A g.q.p. is total iff for any events $A$, $B$, either $A \leq B$ or $B \leq A$.*

- *A g.q.p. is standard iff $A \cap B = \emptyset$ and $A \neq \emptyset$ imply $B < A \cup B$.*

- *A g.q.p. is purely non-standard iff $A < B$ implies $B - A \sim A \cup B$.*

The g.q.p. generated by classical probabilistic models, as described just before Lemma 8, are both total and standard. Standard structures are generated by preference stuctures that satisfy a strengthening of

(Q4), excluding the second possibility in the conclusion. A g.q.p. is standard iff it satisfies the condition: $A \leq B$ implies $\bar{B} \leq \bar{A}$. The g.q.p. generated by the non-standard probabilistic models there are total but not always standard. The g.q.p. generated by total orderings on $S$ as described there are total and purely non-standard. Any purely non-standard g.q.p. is total.

Conjecture: any generalized qualitative plausibility structure is the intersection of all the total g.q.p. that extend it. This conjecture stands, even though such structures are not closed under intersection.

Open problems include the search for a uniform way of generating all generalized qualitative probabilities and the consideration of utility theory on these structures. The framework presented here should allow for a decider to specify only a list of pairs of acts that enter the relations $<_A$, for different $A$'s and this should define a system of preferences satisfying our postulates. How should this system be defined?

## 5.2   EQUIVALENCE OF ACTS THAT TAKE THE SAME VALUES ON EQUIVALENT EVENTS

Let us return to the analogue of Theorem 5.2.1 of Savage [Savage, 1954]: two acts that take the same values on events that are pairwise equivalent necessarily equivalent. Suppose, on some event $A$, acts $f$ and $g$ take only a finite set of values $c_i, i = 0, \ldots, n$. Define, for any such $i$, $F_i$ (resp. $G_i$) to be the event on which act $f$ (resp. $g$) gets value $c_i$. Formally, $F_i \overset{\text{def}}{=} \{s \in A \mid f(s) = c_i$ and $G_i \overset{\text{def}}{=} \{s \in A \mid g(s) = c_i$. Suppose, moreover that, for every $i$, $F_i \sim G_i$. We expect this to imply that $f \sim_A g$. I could not prove that this is implied by (Q1)–(Q6) and I conjecture it is not, but I still lack a counter-example.

But an additional, very natural, postulate implies this conclusion. This additional postulate will be presented now. Notice that it is a natural postulate on preferences that is completely original, in the sense that it does not resemble any of Savage's postulates. Savage's Theorem 5.2.1 requires his Postulate P6, that assumes the existence of fine partitions. No requirement of the sort will be needed here.

(**Q7**)   If $A \cap B = \emptyset$, $A \cup B \subseteq D$, $w_A^{c,d} \leq_{A \cup B} w_B^{c,d}$,

$f =_A d$, $g =_B d$, $f' =_A c$, $g' =_B c$,

$f' =_{D-A} f$, $g' =_{D-B} g$

and $f \leq_D g$ then $f' \leq_D g'$.

The meaning of (Q7) is the following. One concludes $f' \leq_D g'$ because $f \leq_D g$ and $f'$ is very similar to $f$ and $g'$ is very similar to $g$: $f' =_{D-A} f$ and $g' =_{D-B} g$. The difference between $f'$ and $f$ (and $g'$ and $g$) lies in the fact that, where $f$ has value $d$ on $A$, $f'$ has value $c$ there (and where $g$ has value $d$ on $A$, $g'$ has value $c$ there). To simplify things, suppose $d < c$. Then $f'$ is better than $f$ and $g'$ is better than $g$ and the *difference* lies in the difference between $c$ and $d$ and the respective sizes of $A$ and $B$. The assumption that $w_A^{c,d} \leq_{A \cup B} w_B^{c,d}$ implies, given that $d < c$, that $B$ is at least as probable as $A$ (at least as far as $d$ and $c$ are concerned, since we have no postulate implying this this does not depend on the prizes $d$ and $c$). In this case, improving $g$ on $B$ (by going fromn $d$ to $c$) is at least as significant as improving $f$ on $A$ in the same way, and, therefore, $f'$ cannot be preferred to $g'$.

In (Q7), we assumed that $A$ and $B$ were disjoint. One easily sees that this assumption may be dropped: consider $A' \stackrel{\text{def}}{=} A - B$ and $B' \stackrel{\text{def}}{=} B - A$ and then use the Sure Thing Principle, Lemma 5.

With the help of (Q7), we can prove the following analogue to Savage's Theorem 5.2.1.

**Theorem 2** *Assume preferences on acts satisfy (Q1)–(Q7). Let $Z$ be a finite subset of the set $F$ of consequences, $f$ and $g$ acts, and let $A$ be an event. Assume that, for any $s \in A$, both $f(s)$ and $g(s)$ are elements of $Z$. For any $z \in Z$, let $\phi_z \stackrel{\text{def}}{=} \{s \in A \mid f(s) = z\}$ and $\psi_z \stackrel{\text{def}}{=} \{s \in A \mid g(s) = z\}$. Assume that, for any $z \in Z$, $\phi_z \sim \psi_z$. Then, $f \sim_A g$.*

**Proof:** By induction on the size of $Z$. If $Z$ is empty, then $A$ is empty and the claim holds by (Q2). If $Z$ has one element, then $f =_A g$ and the claim holds by (Q2). Let us assume that $Z$ has $n + 1$ elements for $n \geq 1$ and that the claim holds for $Z$ of size $n$.

Let $z_i$, $i = 0, 1$ be distinct elements of $Z$. Define $f'$ to be equal to $f$, except on $\phi_{z_0}$ where it is equal to $z_1$. Define $g'$ to be equal to $g$, except on $\psi_{z_0}$ where it is equal to $z_1$. On $A$, $f'$ and $g'$ take only $n$ different values. Since $\phi_{z_i} \sim \psi_{z_i}$, for $i = 0, 1$, by Lemma 14, we have $\phi_{z_0} \cup \phi_{z_1} \sim \psi_{z_0} \cup \psi_{z_1}$ and the assumptions of Theorem 2 hold and, by the induction hypothesis, $f' \sim_A g'$. The conclusion that $f \sim_A g$ now follows from (Q7) (we noticed above the restriction that the events $A$ and $B$ of (Q7) be disjoint could be removed), since $\phi_{z_0} \sim \psi_{z_0}$ imply

$$w_{\phi_{z_0}}^{z_0,z_1} \leq_{\phi_{z_0} \cup \psi_{z_0}} w_{\psi_{z_0}}^{z_0,z_1}.$$

∎

The results above may contain an avenue to strengthen Savage's results by weakening his (P6). It may be possible to replace, in the proof of Savage's Theorem 5.3.4, his postulate (P6) by the weaker (Q7) and some assumption, similar to Scott's [Scott, 1964], implying that subjective probabilities are defined by probability measures.

## Acknowledgements

Thanks are due to R. Aumann, D. Dubois, H. Prade and D. Schmeidler for their remarks and comments. This work was partially supported by the Jean and Helene Alfassa fund for research in Artificial Intelligence and by grant 136/94-1 of the Israel Science Foundation on "New Perspectives on Nonmonotonic Reasoning".

## References

[Anscombe and Aumann, 1963] F. J. Anscombe and R. J. Aumann. A definition of subjective probability. *Annals of Mathematical Statistics*, 34:199–205, 1963.

[Aumann, 1962] Robert J. Aumann. Utility theory without the completeness axiom. *Econometrica*, 30(3):445–461, July 1962. see also Erratum Vol. 32 pp.210–212, 1964.

[Blackwell and Dubins, 1975] D. Blackwell and L. Dubins. On existence and non-existence of proper, regular conditional distributions. *The Annals of Probability*, 3:741–752, 1975.

[Blume *et al.*, 1991] Lawrence Blume, Adam Brandenburger, and Eddie Dekel. Lexicographic probabilities and choice under uncertainty. *Econometrica*, 59(1):61–79, January 1991.

[de Finetti, 1972] Bruno de Finetti. *Probability, Induction and Statistics*. Wiley, 1972.

[Grätzer, 1971] George Grätzer. *Lattice Theory*. W. H. Freeman, San Francisco, 1971.

[Savage, 1954] Leonard J. Savage. *The Foundations of Statistics*. John Wiley, 1954.

[Scott, 1964] Dana S. Scott. Measurement models and linear inequalities. *Journal of Mathematical Psychology*, 1:233–247, 1964.

[von Neumann and Morgenstern, 1947] John von Neumann and Oskar Morgenstern. *Theory of Games and Economic Behavior*. Princeton University Press, second edition edition, 1947.